\documentclass[11pt]{article}

\usepackage[final]{utils/acl}

\usepackage{times}
\usepackage{latexsym}
\usepackage{float}
\usepackage[T1]{fontenc}

\usepackage[utf8]{inputenc}

\usepackage{microtype}

\usepackage{inconsolata}

\newcommand{\ours}{TAP4LLM}

\usepackage{booktabs} 
\usepackage{amsmath}
\usepackage{amsthm}
\usepackage{amsfonts}       
\usepackage{nicefrac}       
\usepackage{xspace}
\usepackage{xcolor}
\usepackage{backnaur}
\usepackage{multirow}
\usepackage{makecell}
\usepackage{circledsteps}
\usepackage{color}
\usepackage{colortbl}
\usepackage{graphicx}

\usepackage{enumitem}
\setlist[itemize]{align=parleft,left=0pt..1em}
\setenumerate[1]{itemsep=0pt,partopsep=0pt,parsep=\parskip,topsep=5pt}
\setitemize[1]{itemsep=0pt,partopsep=0pt,parsep=\parskip,topsep=5pt}
\setdescription{itemsep=0pt,partopsep=0pt,parsep=\parskip,topsep=5pt}

\makeatletter
\def\verbatim{\small\@verbatim \frenchspacing\@vobeyspaces \@xverbatim}
\makeatother

\makeatletter
\newcommand{\printfnsymbol}[1]{%
  \textsuperscript{\@fnsymbol{#1}}%
} 
\makeatother

\theoremstyle{definition}

\newcommand{\refsec}[1]{\S\ref{#1}} 

\def\eg{\textit{e.g.}\xspace}

\def\etc{\textit{etc}\xspace}
\def\ie{\textit{i.e.}\xspace}
\def\vs{\textit{vs.}\xspace}
\def\wrt{\textit{w.r.t.}\xspace}

\setlist[itemize]{noitemsep}

\usepackage{listings}
\usepackage{xcolor}
\definecolor{codegreen}{rgb}{0,0.6,0}
\definecolor{codegray}{rgb}{0.5,0.5,0.5}
\definecolor{codepurple}{rgb}{0.58,0,0.82}
\definecolor{backcolour}{rgb}{0.95,0.95,0.92}

\lstdefinestyle{mystyle}{
    backgroundcolor=\color{backcolour},   
    commentstyle=\color{codegreen},
    keywordstyle=\color{magenta},
    numberstyle=\tiny\color{codegray},
    stringstyle=\color{codepurple},
    basicstyle=\ttfamily\footnotesize,
    breakatwhitespace=false,         
    breaklines=true,                 
    captionpos=b,                    
    keepspaces=true,                 
    numbers=left,                    
    numbersep=5pt,                  
    showspaces=false,                
    showstringspaces=false,
    showtabs=false,                  
    tabsize=2
}

\title{\ours{}: Table Provider on Sampling, Augmenting, and Packing Semi-structured Data for Large Language Model Reasoning}

\makeatletter
\renewcommand*{\@fnsymbol}[1]{\ensuremath{\ifcase#1\or *\or \dagger\or \ddagger\or
   \mathsection\or \mathparagraph\or \|\or **\or \dagger\dagger
   \or \ddagger\ddagger \else\@ctrerr\fi}}
\makeatother

\author{
Yuan Sui\textsuperscript{\rm 1}\thanks{\indent Equal contribution.}\thanks{\indent The contributions by Yuan Sui, Jiaru Zou, and Xinyi He have been conducted and completed during their internships at Microsoft.}, 
Jiaru Zou\textsuperscript{\rm 2}\printfnsymbol{1}\printfnsymbol{2}, 
Mengyu Zhou\textsuperscript{\rm 3}\thanks{\indent Corresponding author.}, 
Xinyi He\textsuperscript{\rm 4}\printfnsymbol{2},\\
\textbf{Lun Du\textsuperscript{\rm 5}\thanks{\quad The contributions by Lun Du have been conducted and completed when he was a full-time researchers at Microsoft.},}
\textbf{Shi Han\textsuperscript{\rm 3},} 
\textbf{Dongmei Zhang}\textsuperscript{\rm 3} \\
\textsuperscript{\rm 1} National University of Singapore
\textsuperscript{\rm 2} University~of~Illinois~Urbana-Champaign\\
\textsuperscript{\rm 3} Microsoft 
\textsuperscript{\rm 4} Xi'an Jiaotong University 
\textsuperscript{\rm 5} Ant Research
 \\
\texttt{\href{mailto:yuansui@comp.nus.edu.sg}{yuansui@comp.nus.edu.sg}},
\texttt{\href{mailto:jiaruz2@illinois.edu}{jiaruz2@illinois.edu}},
\texttt{\href{mailto:hxyhxy@stu.xjtu.edu.cn}{hxyhxy@stu.xjtu.edu.cn}}\\
\texttt{\{\href{mailto:mezho@microsoft.com}{mezho}, \href{mailto:shihan@microsoft.com}{shihan}, \href{mailto:dongmeiz@microsoft.com}{dongmeiz}\}@microsoft.com},
\texttt{\href{mailto:dulun.dl@antgroup.com}{dulun.dl@antgroup.com}}}

\begin{document}
\maketitle
\begin{abstract}

Table reasoning tasks have shown remarkable progress with the development of large language models (LLMs), which involve interpreting and drawing conclusions from tabular data based on natural language (NL) questions. Existing solutions mainly tested on smaller tables face scalability issues and struggle with complex queries due to incomplete or dispersed data across different table sections.
To alleviate these challenges, we propose \textbf{TAP4LLM} as a versatile pre-processor suite for leveraging LLMs in table-based tasks effectively. It covers several distinct components:
(1) \textit{table sampling} to decompose large tables into manageable sub-tables based on query semantics, (2) \textit{table augmentation} to enhance tables with additional knowledge from external sources or models, and (3) \textit{table packing \& serialization} to convert tables into various formats suitable for LLMs' understanding. In each module, we design and compare several common methods under various usage scenarios, aiming to shed light on the best practices for leveraging LLMs for table-reasoning tasks. Our experiments show that our method improves LLMs' reasoning capabilities in various tabular tasks and enhances the interaction between LLMs and tabular data by employing effective pre-processing.
\end{abstract}

\section{Introduction}
\label{sec:introduction}

The extensive and complex characteristics of the data are commonly represented in the format of structured data. 
\textbf{Table} is one of those fundamental and widely used semi-structured data types in different areas, such as relational databases, spreadsheet applications, and programming languages that handle data for various domains, including financial analysis~\cite{zhang2020a, li2022e}, risk management~\cite{babaev2019}, healthcare analytics~\cite{vamathevan2019}, \etc.
Reasoning over tabular data has several important downstream tasks that are crucial to the field of natural language understanding (NLU) and information retrieval (IR), such as Table-based Question Answering (TQA)~\cite{dataset_hybridqa, dataset_sqa, model_decomposer, model_binder}, Table-based Fact Verification (TFV)~\cite{dataset_tabfact, model_unifiedSKG, model_tableptms}, Table-to-Text~\cite{model_tuta}, Text-to-SQL~\cite{yu2018spider}, Column Type \& Relation~Classification~\cite{model_tabbie, model_turl}, \etc.

\begin{figure}[t]
    \centering
    \includegraphics[width=1\linewidth]{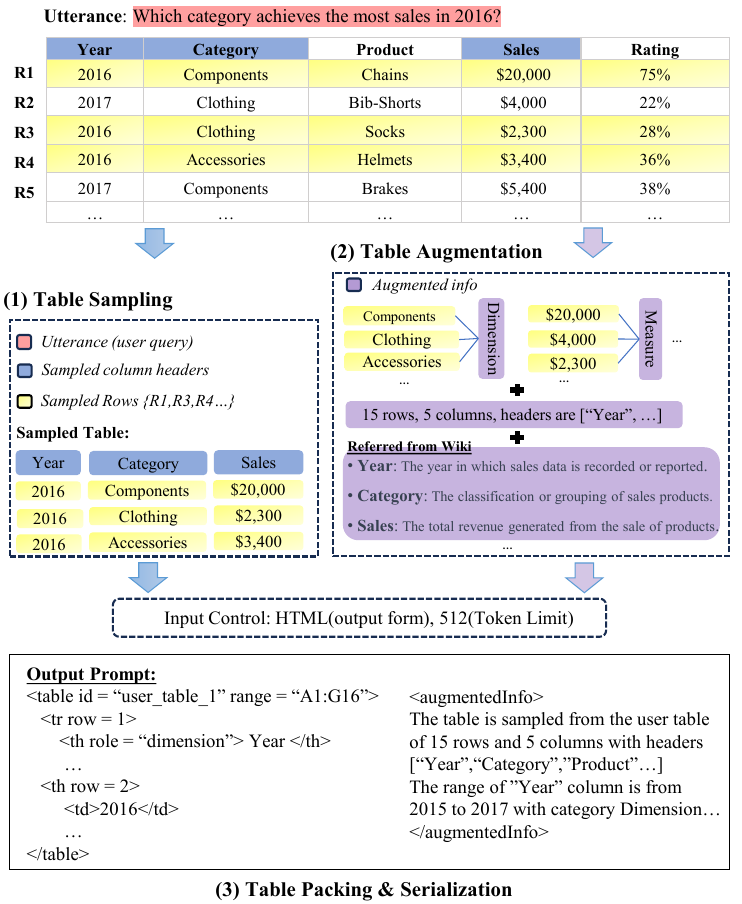}
    \caption{\small{Demo of \ours{} Modules. (1) Table sampling: sample most relevant content. (2) Table augmentation:  retrieve and add extra / meta information. (3) Table packing: serialize the sampled table and augment information into a string while controlling the number of tokens.}}
    \label{fig:illustration}
    \vspace{-3mm}
\end{figure}

Although there is remarkable progress on table reasoning tasks with the development of large language models (LLMs)~\cite{model_binder, model_decomposer, model_tool_synthesis}, the existing solutions are mainly tested on small tables and cannot reflect the real challenges of table reasoning. For example, there are usually issues on scaling to large tables or handling complex queries that require gathering data from various parts of a table. How well do LLMs understand tables and how to leverage LLMs to work with table data remains an open question~\cite{model_table_reasoner, model_tablegpt}. Our research aims to explore this question and shed light on the best practices to leverage LLMs for table-reasoning tasks. Several specific practical issues are faced when leveraging LLMs for table reasoning tasks as follows.

First, which part of a table should be kept in the prompt? The full content of a table could be too lengthy and noisy to be included in the prompt. In addition, most LLMs have a limited input context window size in which an overlong table cannot fit. For large tables that satisfy the length constraint, it can still lead to unnecessary computations (of LLM on long prompt) and quality regressions (generation interfered from noisy input) when placing irrelevant table content (\wrt the task or query) in the prompt.
To address the challenge, some sampling methods were proposed in ad-hoc ways. For example, truncating the input tables to contain only the first 20+ rows and 8 columns~\cite{model_table_reasoner}, or filtering relevant rows based on $n$-gram overlap between them and the utterance~\cite{model_tabert}.
To answer the question of which part to keep, we conduct a systematic study of different grounding and sampling algorithms in Section~\ref{sec:table_sampling}, and the experiments and findings can be found in Section~\ref{sec:exp_table_sampling}.


Second, what additional/external knowledge could help LLMs better understand a table?
The raw content of a table may contain ambiguous information (\eg, abbreviations, domain-specific terms, column type, \etc.) that requires further interpretation and clarification. As a result, direct reasoning with the raw table may lead to misinterpretation and hallucination by LLMs. 
To address this, some augmentation techniques were proposed to incorporate structured knowledge~\cite{model_structure_LLM,model_unifiedSKG}, common sense knowledge~\cite{model_chatgpt_commonsense,model_chatgpt_analysis,model_ischatgpt_trust,model_how_chatgpt_close_human}, and analytical knowledge~\cite{model_table_recasting,he2023anameta} into training and inference processes.
For example, \cite{model_table_recasting} transforms existing tabular data to create diverse NL inference instances for better zero-shot performance. 
AnaMeta~\cite{he2023anameta} infers implicit metadata behind raw table contents through field distribution and knowledge graph information.
However, the techniques were proposed independently and there lack a comprehensive study that compares them and attempts to combine them to provide useful and diverse knowledge and thoughts for LLMs.
We will discuss several augmentation methods in Section~\ref{sec:table_augmentation} and their corresponding experiments and findings can be found in Section~\ref{sec:exp_table_augmentation}.

Moreover, table augmentation plays a crucial role in avoiding LLMs to partitionally comprehend the semantics and distribution of the whole table solely based on table sampling (which may remove some essential rows/columns due to the limitations of the methods). It leverages the summarization, statistics, and metadata information derived from the whole table to represent high-level information, to compromise the trade-off with table sampling, which will intuitively decrease information entropy. We will discuss this essential trade-off in Section~\ref{sec:trade_off_between_token_allocation} and the experiments and findings can be found in Figure~\ref{fig:token-allocation}.


Third, how do we encode the table into a prompt? While sampling and grounding compress the table content, augmentation expands the prompt by adding more information. With a given token budget, one needs to find the balance to allocate available tokens between table content and augmented knowledge. Furthermore, the serialization format of the table also plays a critical role. It not only influences how well an LLM understands the table input~\cite{model_structure_LLM}, but also determines the string length of the serialized table and the augmented information.
For example, as discussed in \cite{model_structure_LLM}, table formats such as HTML~\cite{model_htlm} or XML are better understood by GPT models, but they also lead to increased token consumption. To pack a table into the prompt, these problems should be addressed with trade-offs (see Section~\ref{sec:trade_off_between_token_allocation}).

\begin{figure*}[ht]
    \centering
    \includegraphics[width=1\linewidth]{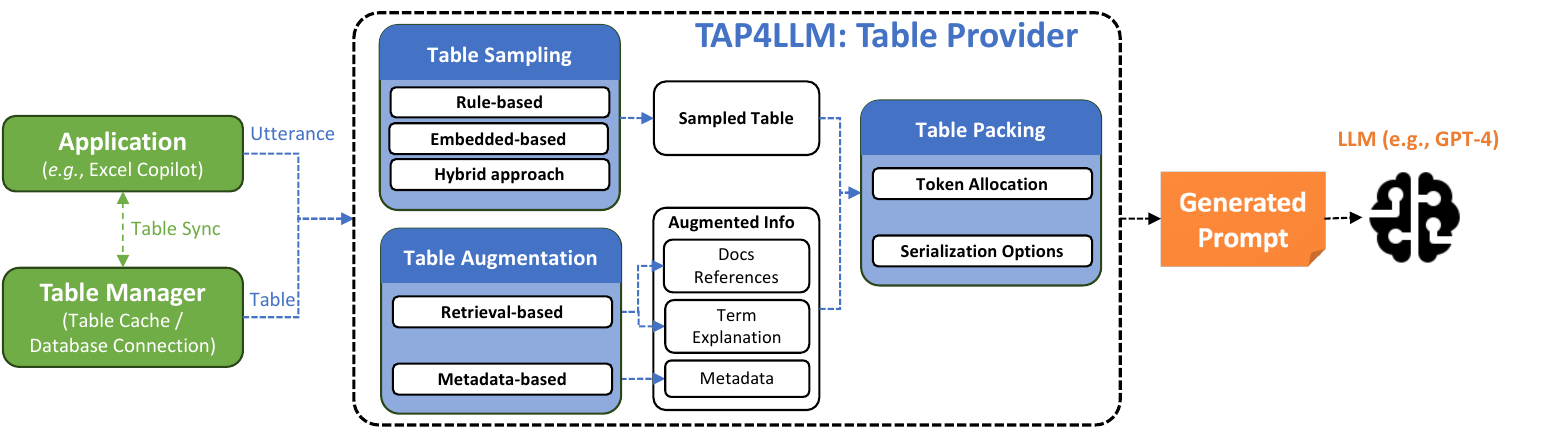}
    \caption{\small{\ours{} Framework for Tabular Data. Note that ``table sync'' refers to the application (such as Excel Copilot) keeping its table data in sync with the table manager. The table manager acts as an intermediary, managing the data that is either stored locally in a cache or accessed through a database connection. This sync process is crucial for ``interactive table reasoning'' and for maintaining data integrity. The implications of this syncing process are further discussed in \refsec{sec:interactive_table_reasoning}.}}
    \label{fig:table_provider_framework}
\end{figure*}

To mitigate the aforementioned challenges, we propose \textbf{\ours{}} (\textbf{TA}ble \textbf{P}rovider for \textbf{L}arge \textbf{L}anguage \textbf{M}odels) as a versatile pre-processor suite for LLMs in table-based tasks. Specifically, there are three essential modules:
(\romannumeral1) \underline{Table Sampling}: decomposing large tables into manageable sub-tables based on query semantics;
(\romannumeral2) \underline{Table Augmentation}: enhancing tables with additional knowledge from external sources or symbolic models; and 
(\romannumeral3) \underline{Table Packing}: convert tables into various formats (\textit{e.g.}, HTML, XML, Markdown, \textit{etc}) suitable for LLMs' understanding while balancing the token allocation trade-off as well.

In each module, we design and compare new and existing methods for various scenarios across six distinct datasets. Through experiments in Section~\ref{sec:experiments}, we find that:
(1) When using LLMs to process tables, it is more effective to concentrate on key rows and columns rather than overloading with extraneous data. For tasks that require high accuracy, semantic-based sampling typically enhances performance, whereas for tasks prioritizing low latency and minimal computational resources, rule-based sampling may be more suitable.
(2) Integrating external knowledge of the tables can consistently improve the performance of table reasoning tasks by reducing hallucinations and factual inaccuracies in LLMs and improving general comprehension and analysis of tabular data.
(3) A balanced distribution of tokens between table content and augmented information can help improve overall performance; We find a performance-optimized ratio (close to 5:5 or 4:6 between table content and augmentation tokens) often achieves the best performance across different settings.

In summary, our \textbf{main contributions} are:
\begin{itemize}[noitemsep, left=0pt]
    \item We propose \ours{} framework to improve the effectiveness of LLMs on tabular reasoning tasks by better sampling, augmentation and packing tables into input prompts.
    
    \item A comprehensive evaluation of each component is conducted. On average \ours{} achieves a 7.93\% performance improvement. 
    
    \item We summarize a table prompting guideline. For different real-world scenarios, we identify the corresponding optimal combination / setting of methods within each module.
\end{itemize}



\section{\ours{}: Table Provider for LLMs} 
\label{sec:method}

The overall architecture of \ours{} is defined as follows (as illustrated in Figure~\ref{fig:table_provider_framework}): 
Given a natural language query/utterance $Q$ from applications (\eg, Excel Copilot) and a table $T$ from Table Manager (\eg, table cache or database connection), our system incorporates three core components as follows:
\begin{itemize}[noitemsep, left=0pt]
    \item \textbf{Table~Sampling}: Decompose a large table $T$ into a sub-table $T'$ with specific rows and columns.
    \item \textbf{Table~Augmentation}: Incorporate relevant external knowledge, metadata, and attributes about the original table $T$ explicitly.
    \item \textbf{Table~Packing}: Convert tables into various formats suitable for LLMs’ understanding while control the token allocation for table sampling and augmentation.
\end{itemize}

\subsection{Table Sampling}
\label{sec:table_sampling}

In table sampling, a subset of top-ranked rows and columns is selected to form a sub-table. Specifically, given an original table $T$ with a distinct set of rows $R_T$, columns $C_T$, and a query $q$, the goal of table sampling is to produce a sub-table $T' = T_{r,c}$, where $r \in \mathcal{P}(R_T)$, $c \in \mathcal{P}(C_T)$. Here $\mathcal{P}(X)$ denotes the power set of $X$, representing all possible subsets of $X$. The process can be formulated as 
\begin{equation}
    T' = T_{r,c} = select(T, rank(f(T, q)))
    \label{Eq(1)}
\end{equation}
The $f(T,q)$ function represents each sampling method. For example, the query-based sampling (discussed in detail below) calculates the similarity score as $f$ between the query $q$ and each row/column from $T$. The $rank()$ function sorts the rows and columns of $T$ based on sampling methods $f$ and outputs a ranked list. The $select()$ function then chooses the top-k rows and top-l columns from the ranked list to form the sub-table $T_{r,c}$. 
Specifically, we classify multiple variants for table sampling as three following categories:

\subsubsection{Rule-based Sampling}
\label{method_sampling:rule-based-sampling} 
Rule-based sampling refers to table sampling based on predefined criteria or rules. These methods follow the established patterns or criteria for data selection. Specifically, we consider three common rule-based sampling methods incoporated into our table sampling module: (1) \textit{Random Sampling}, (2) \textit{Evenly Sampling}, and (3) \textit{Content Snapshot \& Synthetically Sampling}. The detailed description can be found in Appendix~\ref{sec:rule-based-sampling}.

\subsubsection{Embedding-based Sampling}
\label{method_sampling:embedding-based-sampling}
Instead of adhering to strict rules or criteria in rule-based sampling, embedding-based methods leverage the semantic and contextual representation of each row and column. Specifically, let $T$ be a table where $R_T$ is the set of rows and $C_T$ is the set of columns. Let $E: R_T \cup C_T \rightarrow \mathbb{R}^d$ be an embedding function that maps each row or column to a $d$-dimensional vector by capturing its semantic content. By mapping each row or column to vectors, this method harnesses the power of spatial relationships within the embedding space to guide sampling decisions. 

Here, we propose three variant methods as shown below:
(1) \textit{Semantic-based Sampling}:
Semantic-based Sampling is a tailored approach emphasizing the semantics relevance of row/columns to the utterance. The process is exactly illustrated in Eq.~\ref{Eq(1)}. Note that the default query-based sampling is the row-based method. In our experiments, we also study the column grounding shown in Table~\ref{tab:table_sampling}. 
(2) \textit{Centroid-based Sampling}:
The goal of centroid-based sampling is to ensure the preservation of data diversity. 
We use \textit{K-Means}~\cite{macqueen1967some} to partition the set of embeddings into $n$ clusters $C_n$. For each cluster $C_i$, we select the top-$K$ rows or columns based on the closeness to the centroid. 
(3) \textit{Hybrid-approach}:
The Hybrid approach marries the specificity of semantic-based sampling with the broad representations of centroid-based sampling. Specifically, the top-$K$ rows or columns are selected based on a combination metric $h(r, c, u)$ measuring the directional distance to cluster centroid $c$ and the semantic similarity to the utterance $u$. We formulate the measuring metric as:
\begin{equation}
    h(r, c, u)=\alpha(\frac{1}{1+D(r,c)}) + \beta S(r,u)
\end{equation}
where $D(r,c)$ measures the directional distance (\eg, Euclidean distance) between selected rows or columns and cluster centroid in embedding space, and $S(r,u)$ measures the semantic similarity between rows/columns and the utterance. 
The weights $\alpha$ and $\beta$ provide flexibility in prioritizing between contextual relevance and diversity. 
Without further specification, we set $\alpha=0.3$ and $\beta=0.7$ in our experiments.

\subsubsection{LLM-based Sampling}
LLMs have proven effective in tabular reasoning~\cite{model_decomposer}, utilizing their capabilities to predict row and column indices for efficient sub-table extraction.
However, relying on LLMs for pre-processing significantly increases computational costs. Additionally, using LLMs to predict indices introduces challenges such as token limitations, noisy information, and the need for further table pre-processing, turning the task into a recursive loop.
Despite this method not being ideally suited to our scope, we still consider it a strong baseline, albeit at the expense of time.

\subsection{Table Augmentation}
\label{sec:table_augmentation}
Table augmentation enhances LLM reasoning by adding extra knowledge to the input table and query.
In the table augmentation module, we categorize different knowledge aspects into three main categories: Metadata-based, Retrieval-based, and Self-consistency-based (See Table \ref{tab:table_augmentation_definition} in Appendix). We will describe each category in detail below. 

\subsubsection{Metadata-based Augmentation}
\label{method_augmentation:symbolic_agents}
Tabular data analysis relies on accurately understanding field semantics and identifying common patterns in everyday analysis~\cite{he2023anameta}. 
Following AnaMeta~\cite{he2023anameta} using a range of knowledge-fusion language models for metadata inference, we consider five main metadata-based augmentation types and leverage LLMs for zero-shot inference using metadata instruction as clues: \textit{Dimension / Measure}, \textit{Semantic Field Type},  \textit{Table Size}, \textit{Statistics Feature}, \textit{Header Hierarchy}. The detailed description for each meta-data augmentation type can be found in Appendix~\ref{apx:metadata_type}.

\begin{table*}[htbp]
  \centering
  \caption{\small{Comparative results of the table sampling methods. The term ``w/ Column Grounding'' refers to the method consider both row-based and column-based sampling (sometimes referred to as ``grounding''). ``GPT-3.5'' refers to the OpenAI released model gpt-3.5-turbo-32k, with 32k token-sized context window; In contrast, ``GPT-3.5 truncated'' refers to gpt-3.5-turbo, with 4k token-sized context window, where most tables will be truncated according to this token limitation. The top-3 performances on each dataset are highlighted in green, with the best performance being both bold and underlined.}
}
  \label{tab:table_sampling}
  \resizebox{\linewidth}{!}{
    \begin{tabular}{llccccc}
    \toprule
     Sampling Type & \multicolumn{1}{l}{Table Sampling Methods} & SQA & FEVEROUS & TabFact  & HybridQA & ToTTo\\
    \midrule
    \multirow{3}{*}{Rule-based Sampling} & Random Sampling & 27.30\% & 60.30\% & 55.17\% & 23.60\%& 40.12\% \\
    & Evenly Sampling & 26.72\% & 61.87\% & 54.63\% &5.32\% &  29.41\% \\
    & Content Snapshot~\cite{model_tabert} & 28.24\% & 63.10\%  & 56.92\% &23.40\%  &  47.51\%\\
    \midrule
    \multirow{4}{*}{Embedding-based Sampling} &
    Centroid-based Sampling & 28.10\%  & \cellcolor[rgb]{ .776,  .937,  .808}\textcolor[rgb]{ 0,  .38,  0}{63.50\%} & 55.40\% &  24.03\% & 48.30\% \\
    
    & Semantic-based Sampling & \cellcolor[rgb]{ .776,  .937,  .808}\textcolor[rgb]{ 0,  .38,  0}{28.32\%} & 63.32\% & 59.80\% &   24.32\% & \cellcolor[rgb]{ .776,  .937,  .808}\textcolor[rgb]{ 0,  .38,  0}{49.14\%} \\
    
    & \quad w/ Column Grounding & \cellcolor[rgb]{ .776,  .937,  .808}\textcolor[rgb]{ 0,  .38,  0}{\underline{\textbf{29.12\%}}} & \cellcolor[rgb]{ .776,  .937,  .808}\textcolor[rgb]{ 0,  .38,  0}{64.74\%} & \cellcolor[rgb]{ .776,  .937,  .808}\textcolor[rgb]{ 0,  .38,  0}{60.23\%} &  \cellcolor[rgb]{ .776,  .937,  .808}\textcolor[rgb]{ 0,  .38,  0}{\underline{\textbf{25.14\%}}}& \cellcolor[rgb]{ .776,  .937,  .808}\textcolor[rgb]{ 0,  .38,  0}{\underline{\textbf{53.42\%}}} \\
    
    & Hybrid Sampling & \cellcolor[rgb]{ .776,  .937,  .808}\textcolor[rgb]{ 0,  .38,  0}{28.79\%} & \cellcolor[rgb]{ .776,  .937,  .808}\textcolor[rgb]{ 0,  .38,  0}{\underline{\textbf{65.34\%}}} & \cellcolor[rgb]{ .776,  .937,  .808}\textcolor[rgb]{ 0,  .38,  0}{\underline{\textbf{61.37\%}}} &  \cellcolor[rgb]{ .776,  .937,  .808}\textcolor[rgb]{ 0,  .38,  0}{24.71\%}& \cellcolor[rgb]{ .776,  .937,  .808}\textcolor[rgb]{ 0,  .38,  0}{51.63\%} \\
    \midrule
    \multirow{1}{*}{LLM-based Sampling} & LLM-Decomposer~\cite{ye2023} & 27.98\% & 62.34\% & \cellcolor[rgb]{ .776,  .937,  .808}\textcolor[rgb]{ 0,  .38,  0}{58.74\%} & \cellcolor[rgb]{ .776,  .937,  .808}\textcolor[rgb]{ 0,  .38,  0}{24.98\%} &  48.13\% \\
    \midrule
    \multirow{2}{*}{-} & No sampling (GPT-3.5) & 27.60\% & 60.12\% & 56.20\% &  14.10\% &47.42\%\\
    
    & No sampling (GPT-3.5, truncated) & 23.54\% & 43.54\% & 52.12\% & 23.12\% &30.42\% \\
    
    
    \bottomrule
    \end{tabular}
    }
\end{table*}%

\subsubsection{Retrieval-based Augmentation}
\label{method_augmentation:retrieval_based_approach}

Large Language Models have occasionally been observed to generate hallucinated or factually inaccurate text~\cite{hallucinate, LLMsurvey}. To mitigate these issues, several works have proposed to strengthen LLMs with information retrieval systems~\cite{RAG1, RAG2, WebGPT}, which enables LLMs to retrieve relevant content from an external repository (knowledge corpus). It has been verified that retrieval-augmented LLMs can generate texts in response to user input with fewer hallucinations~\cite{WebGPT}. Furthermore, by incorporating customized private data resources, retrieval-augmented LLMs can respond to in-domain queries that cannot be answered by LLMs trained with public data. As previous works~\cite{WebGPT, RAG1, RAG2} suggested, LLMs can generate more factual answers by feeding the references retrieved from the external corpus.

In \ours{}, we have fortified the document retrieval capabilities of LLMs and consider two components: (1) \textit{document references}: provide supplemental relevant web pages as the references for the given table; (2) \textit{term explanation}: explain strange/ambiguous term in the given table. We utilize technologies including vector databases~\cite{wang2021vecdb} and LangChain~\cite{langchain} to facilitate the retrieval of pertinent information from Wikipedia\footnote{\url{https://www.wikipedia.org/}}. The details for document references and term explanation can be found in Appendix~\ref{sec:retrieval-based-augmentation}.

\subsubsection{Self-consistency-based Augmentation}
\label{method_augmentation:self_consistency_based_approach}
We follow \cite{model_structure_LLM} to implement the self-consistency-based augmentation approach. Specifically, we append the instruction \textit{"Identify critical values and ranges of the last table related to the statement"} to the initial prompt and forward it to the LLM. The output generated from this instruction is then incorporated back into the prompt.
Following this, we then re-forward the enriched prompt to LLMs — containing both the initial query and the newly generated insights, along with task-specific instructions for further processing.

\subsection{Table Packing}
\label{sec:table_packing_serialization}
The table packing module is motivated by the need to preserve efficient reasoning without altering the LLM architecture.
First, we apply this module to manage token-limit allocation for the table sampling and augmentation. To achieve this, we conduct an empirical study to determine the optimal ratio between the sub-table length and the length of the augmentation information, as illustrated in Figure~\ref{fig:token-allocation}. The packing process is regulated by a user-defined token limit parameter, which sets the maximum truncated token length.
Additionally, we are inspired by the study~\cite{model_structure_LLM}, which highlights that using markup languages such as \textit{HTML} or \textit{XML} significantly improves generation quality in comparison to TQA and TFV. In line with this, \ours{} supports multiple serialization formats, including HTML, XML, JSON, CSV, NL+Sep (a common option, \eg, using `$|$' as a cell/column separator), and Markdown, \etc.

\begin{table*}[t]
  \centering
  \caption{\small{Comparative results of the table augmentation methods. We use a semantic-based sampling method without augmentation as the baseline. The term ``Delta'' refers to the performance gap between each method and the baseline. The top-3 performance gaps on each dataset are highlighted in green, with the best performance being underlined. Note that since only the ToTTo dataset contains hierarchical headers, we only provide the ``header hierarchy'' method on this dataset. ``D/M + SF" refers to Dimension/Measure+ Semantic Field Type.}}
  \label{tab:table_augmentation}
  \resizebox{\textwidth}{!}{
    \begin{tabular}{lcccccccccc}
    \toprule
    \multirow{2}[4]{*}{\textbf{Augmentation Aspect}} & \multicolumn{2}{c}{SQA} & \multicolumn{2}{c}{FEVEROUS} & \multicolumn{2}{c}{TabFact} & \multicolumn{2}{c}{HybridQA} & \multicolumn{2}{c}{ToTTo}\\
    \cmidrule{2-11}      & Acc   & Delta & Acc   & Delta & Acc   & Delta & Acc   & Delta & BLEU-4 & Delta\\
    \midrule
    baseline & 28.32\% & 0.00\% & 63.32\% & 0.00\% & 59.80\% & 0.00\% & 24.32\% & 0.00\% & 49.14\% & 0.00\% \\
    \midrule
    D/M + SF & 30.12\% & 1.80\% & 65.72\% & \cellcolor[rgb]{ .776,  .937,  .808}\textcolor[rgb]{ 0,  .38,  0}{2.40\%} & {\textbf{62.67\%}} & \cellcolor[rgb]{ .776,  .937,  .808}\textcolor[rgb]{ 0,  .38,  0}{\underline{2.87\%}} & 26.12\% & 1.80\% & 51.25\% & 2.11\% \\
    Table Size & 28.85\% & 0.53\% & 63.40\% & 0.08\% & 60.30\% & 0.50\% & 24.94\% & 0.62\% & 49.03\% & -0.11\% \\
    Statistics Feature & 31.22\% & \cellcolor[rgb]{ .776,  .937,  .808}\textcolor[rgb]{ 0,  .38,  0}{2.90\%} & {\textbf{66.51\%}} & \cellcolor[rgb]{ .776,  .937,  .808}\textcolor[rgb]{ 0,  .38,  0}{\underline{3.19\%}} & 62.33\% & \cellcolor[rgb]{ .776,  .937,  .808}\textcolor[rgb]{ 0,  .38,  0}{2.53\%} & 26.13\% & \cellcolor[rgb]{ .776,  .937,  .808}\textcolor[rgb]{ 0,  .38,  0}{1.81\%} & 50.57\% & 1.43\%\\
    Header Hierarchy & -     & -     & -     & -     & -     & -     & -     & -     & 48.64\% & -0.50\% \\
    \midrule
    Docs References & {\textbf{33.45\%}} & \cellcolor[rgb]{ .776,  .937,  .808}\textcolor[rgb]{ 0,  .38,  0}{\underline{5.13\%}} & 63.13\% & -0.19\% & 61.32\% & 1.52\% & 25.12\% & 0.80\% & 52.74\% & \cellcolor[rgb]{ .776,  .937,  .808}\textcolor[rgb]{ 0,  .38,  0}{3.60\%}\\
    Term Explanations & & & & & & & & & &\\
    \quad - LLM-based & 31.59\% & \cellcolor[rgb]{ .776,  .937,  .808}\textcolor[rgb]{ 0,  .38,  0}{3.27\%} & 64.12\% & 0.80\% & 62.32\% & \cellcolor[rgb]{ .776,  .937,  .808}\textcolor[rgb]{ 0,  .38,  0}{2.52\%} & 26.24\% & \cellcolor[rgb]{ .776,  .937,  .808}\textcolor[rgb]{ 0,  .38,  0}{1.92\%} & {\textbf{53.21\%}} & \cellcolor[rgb]{ .776,  .937,  .808}\textcolor[rgb]{ 0,  .38,  0}{\underline{4.07\%}} \\
    \quad - Heuristics-based & 29.59\% &1.27\% & 63.72\% &0.40\% & 61.58\% &1.78\% & 25.24\%& 0.92\% & 51.21\% & 2.07\%\\
    \midrule
    Self Prompting & 30.45\% & 2.13\% & 65.24\% & \cellcolor[rgb]{ .776,  .937,  .808}\textcolor[rgb]{ 0,  .38,  0}{1.92\%} & 62.32\% & \cellcolor[rgb]{ .776,  .937,  .808}\textcolor[rgb]{ 0,  .38,  0}{2.52\%} & {\textbf{26.64\%}} & \cellcolor[rgb]{ .776,  .937,  .808}\textcolor[rgb]{ 0,  .38,  0}{\underline{2.32\%}} & 52.36\% & \cellcolor[rgb]{ .776,  .937,  .808}\textcolor[rgb]{ 0,  .38,  0}{3.22\%}\\
    \bottomrule
    \end{tabular}%
  }
\end{table*}

\section{Experiments}
\label{sec:experiments}
In this section, we first present the experimental setup, followed by an extensive comparison between baselines within each module in \ours{}. Additionally, we conduct an ablation study and provide a thorough evaluation of the performance of \ours{}. For further details on the experimental setup and additional experiments, please refer to Appendix~\ref{sec:experiment_settings} and~\ref{sec: additional_experiments}.

\subsection{Experiment Settings}

\textbf{Datasets.}
We evaluate \ours{} on five TQA \& TFV datasets: Sequential Question Answering (SQA)~\cite{dataset_sqa}, HybridQA~\cite{dataset_hybridqa}, TabFact~\cite{dataset_tabfact}, ToTTo~\cite{dataset_totto}. Additionally, we set up \ours{} on a Text-to-SQL database Spider~\cite{yu2018spider}, detailed in ~\ref{sec: spider}. The statistics of the datasets are given in Table~\ref{tab:dataset_distribution}, and the details of the datasets and metrics are described in Appendix~\ref{sec:tasks_and_datasets}.


\noindent\textbf{Models.}
We select state-of-the-art LLMs that have been widely studied in text generation and reasoning, including multiple GPT-series models and advanced open-source LLMs. Details of the tested models and embedding methods are provided in Appendix~\ref{sec:models}, while the experimental results for open-source LLMs are available in Appendix~\ref{sec: open_source}.

\subsection{Results of Table Sampling}
\label{sec:exp_table_sampling}
As shown in Table~\ref{tab:table_sampling}, we perform comparative experiments on various table sampling methods, leading to the following key observations:
(1) \textit{Semantic-based sampling} with column grounding outperforms other sampling methods across all datasets by effectively selecting table parts most relevant to queries. \textit{Centroid-based sampling} also shows competitive results by clustering data points within tables, though it lacks query-table relevance consideration. Moreover, when combining these two strong variants (\textit{hybrid sampling}), it shows the most powerful capability.
(2) The \textit{rule-based sampling} method \textit{content snapshot}, while not as precise in capturing query-specific information, offers a promising, efficient alternative by focusing on essential table content through $n$-gram overlap, without the need for complex embedding calculations. 
(3) Direct encoding methods, including using GPT-3.5-turbo with a 32k token limit or a 4k token-sized context window with truncation, demonstrate inferior performance. This suggests that while they can encompass more table information, they may introduce noise or lose critical context, undermining the table reasoning process and highlighting the importance of strategic sampling for optimal LLM performance.

\subsection{Results of Table~Augmentation} 
\label{sec:exp_table_augmentation}
For the comparative experiments of table augmentation methods, we use the semantic-based sampling method as the baseline and report the performance gap between adding each augmentation method or not.
Table~\ref{tab:table_augmentation} provides several key insights, summarized as follows:
(1) Table augmentation methods further improve LLM's reasoning ability after sampling. For example, \textit{D/M + SF} achieves higher accuracy across all six datasets (a most significant increase on TabFact +2.87\%). \textit{Docs References} and \textit{Term Explanations} add meaningful context and semantic understanding to the model's processing of tables, with (SQA +5.13\%, ToTTo +4.07\% ). The \textit{Self-Prompting} further exemplifies the potential for iterative improvement in query and response generation. However, not all augmentation methods yield positive outcomes. \textit{Table Size} offers minimal performance enhancement and \textit{Header Hierarchy} shows that introducing a hierarchy may complicate the model's ability to process the tabular information in some contexts, possibly by adding unnecessary complexity.
(2) Additionally, the comparison of cell selection methods for \textit{Term Explanations} highlights the superior performance of LLM-based selection over heuristic approaches. 
We find that LLM-based cell selection outperforms the heuristics-based cell selection with improvements in ``Delta'' ranging from 0.80\% to 4.07\%. 
While achieving higher performance, the LLM-based method also increases the calling budget as it requires additional LLM calls. These results indicate that the method's effectiveness varies with the dataset. i.e. It's beneficial for datasets requiring complex text understanding and generation (SQA and ToTTo). However, its impact is less distinct or even slightly negative in datasets involving different types of data or nuanced tasks (FEVEROUS and HybridQA).

\begin{figure}[htbp]
\vspace{-2mm}
    \centering\includegraphics[width=0.9\linewidth]{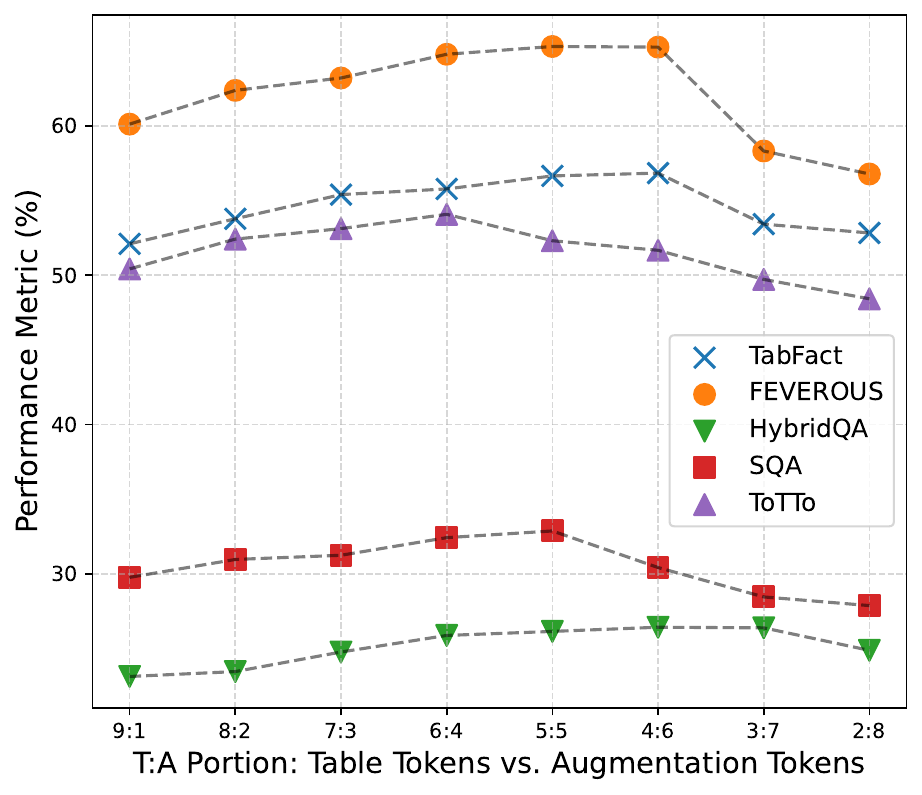}
    \vspace{-2mm}
    \caption{\small{Token Allocation. T:A refers to the ratio of upper \#token limitations of sampled table \vs augment info.}}
    \vspace{-3mm}
    \label{fig:token-allocation}
\end{figure}

Through the experiment results, we also observe that different augmentation methods perform well on the same dataset. For example, \textit{D/M + SF}, \textit{Statistics Feature}, \textit{Term Explanation} and \textit{Self Prompting} all show significant improvement on the TabFact dataset. 
This suggests that combining multiple augmentation methods may have cumulative effects, leading to improved performance. In this work, we only report one simple intuitive approach - appending all augmentation information together into the prompt. We leave more fine-grained combinations for future exploration.

\subsection{Ablation Study of \ours{}}
As shown in Table~\ref{tab:ablation_study}, we conduct an ablation study to evaluate the impact of various components on the performance of \ours{}. Each row represents the model's performance with the removal of a specific component. Our findings indicate that all components contribute to the overall effectiveness of the model, with certain components, such as table sampling and table augmentation, being particularly critical. The study also reveals that each dataset responds differently to the removal of features, underscoring the importance of a tailored design when optimizing for specific datasets.
We also report the performance using the most optimal combination of table sampling and augmentation for each dataset, as presented in Table~\ref{tab:ablation_study}.


\begin{table*}[htbp]
  \centering
  \caption{\small{Ablation results on five table datasets using gpt-3.5-turbo model. Similar to Table 2, the lowest accuracy on each dataset is bold. The top-3 decreasing gap (delta) on each dataset are highlighted in red, with the lowest performance being underlined. The performance of golden combination of table sampling and augmentation (``hybrid-sampling + all-augmentation'') is reported in the first row.}}
  \resizebox{\textwidth}{!}{
    \begin{tabular}{lcccccccccc}
    \toprule
    \multirow{2}[4]{*}{Components of TAP4LLM} & \multicolumn{2}{c}{SQA} & \multicolumn{2}{c}{FEVEROUS} & \multicolumn{2}{c}{TabFact} & \multicolumn{2}{c}{HybridQA} & \multicolumn{2}{c}{ToTTo} \\
    \cmidrule{2-11} & Acc   & Delta & Acc   & Delta & Acc   & Delta & Acc   & Delta & BLEU-4 & Delta \\
    \midrule
    All & 34.12\%  & 0.00\%& 68.32\%  & 0.00\%&64.78\% & 0.00\%& 27.87\%  & 0.00\%& 54.93\%  & 0.00\%\\
    \midrule
    w/o table sampling & \textbf{26.54\%}  & \cellcolor[rgb]{1, 0.6, 0.6}\textcolor[rgb]{0.4, 0, 0}{\underline{-7.58\%}} & \textbf{61.54\%} &\cellcolor[rgb]{1, 0.6, 0.6}\textcolor[rgb]{0.4, 0, 0}{\underline{-6.78\%}} & \textbf{58.12\%} &\cellcolor[rgb]{1, 0.6, 0.6}\textcolor[rgb]{0.4, 0, 0}{\underline{-6.66\%}} & \textbf{24.12\%} &\cellcolor[rgb]{1, 0.6, 0.6}\textcolor[rgb]{0.4, 0, 0}{\underline{-3.75\%}} & \textbf{48.47\%} & \cellcolor[rgb]{1, 0.6, 0.6}\textcolor[rgb]{0.4, 0, 0}{\underline{-6.46\%}} \\
    w/o table augmentation - all & 29.12\% &\cellcolor[rgb]{1, 0.6, 0.6}\textcolor[rgb]{0.4, 0, 0}{-5.00\%} & 63.74\% &\cellcolor[rgb]{1, 0.6, 0.6}\textcolor[rgb]{0.4, 0, 0}{-4.58\%} & 60.23\% &\cellcolor[rgb]{1, 0.6, 0.6}\textcolor[rgb]{0.4, 0, 0}{-4.55\%} & 25.14\% &\cellcolor[rgb]{1, 0.6, 0.6}\textcolor[rgb]{0.4, 0, 0}{-2.73\%} & 53.42\% & -1.51\% \\
    w/o table augmentation - metadata-based & 33.87\% &-0.25\% & 64.38\% &\cellcolor[rgb]{1, 0.6, 0.6}\textcolor[rgb]{0.4, 0, 0}{-3.94}\% & 62.78\% & \cellcolor[rgb]{1, 0.6, 0.6}\textcolor[rgb]{0.4, 0, 0}{-2.00\%} & 26.98\% & -0.89\% & 53.42\% & -1.51\% \\
    w/o table augmentation - retrieval-based & 31.42\% &\cellcolor[rgb]{1, 0.6, 0.6}\textcolor[rgb]{0.4, 0, 0}{-2.7}\% & 66.23\% & -2.09\% & 62.97\% & -1.81\% & 26.33\% & -1.54\%  & 52.67\% &\cellcolor[rgb]{1, 0.6, 0.6}\textcolor[rgb]{0.4, 0, 0}{-2.26\%} \\
    w/o table packing & 31.87\% & -2.25\%  & 67.42\% & -0.90\% & 63.28\% & -1.50\% & 26.32\% & \cellcolor[rgb]{1, 0.6, 0.6}\textcolor[rgb]{0.4, 0, 0}{-1.55\%}  & 52.87\% & \cellcolor[rgb]{1, 0.6, 0.6}\textcolor[rgb]{0.4, 0, 0}{-2.06\%} \\
    \bottomrule
    \end{tabular}%
  }
  \label{tab:ablation_study}%
\end{table*}


\subsection{Trade-offs between Token Allocation}
\label{sec:trade_off_between_token_allocation}
We employ five table datasets to explore the trade-off between token allocation for table sampling and table augmentation, as illustrated in Figure~\ref{fig:token-allocation}. We observe that:
(1) A balanced token distribution between the table and augmentation (approximately 5:5 or 4:6, referred to as the \textit{balanced T:A ratio}) generally achieves the best performance across all five datasets. This suggests that carefully managing token allocation can enhance LLM performance. 
(2) Diminishing returns are observed when an excessive number of tokens are allocated to augmentation information (e.g., a 3:7 ratio), leading to a decline in performance. This indicates that beyond a certain point, additional augmentation tokens may no longer be beneficial and could detract from the core table content.

The trade-off we analyze above reflects a broader principle in data processing and machine learning: the balance between information overload and information scarcity. Over-augmentation can introduce noise, making it harder to identify key patterns or insights, while excessive sampling may lead to an incomplete or biased understanding of the data. It is important to note that the optimal T:A ratio may vary across datasets, as each has unique characteristics that make certain ratios more effective.

\begin{figure*}[htbp]
    \centering
    \includegraphics[width=1\linewidth]{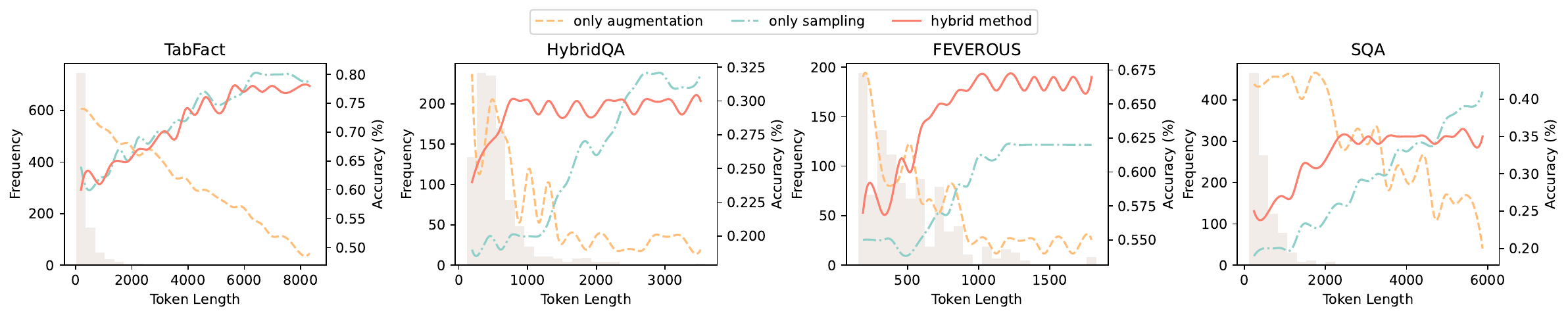}
    \caption{\small{Comparative Analysis of Model Performance Across TabFact, HybridQA, FEVEROUS and SQA.
    The series of graphs illustrates the frequency distribution of token lengths alongside the LLM performance (\%) for three distinct methods: only sampling, only augmentation, and the hybrid method. Each subplot corresponds to a different dataset, depicting how table token length impacts model accuracy for various data augmentation and sampling strategies. Note that the ``only augmentation'' method refers to adding only the augmentation information to the prompt, without using any sampling method.}}
    \label{fig:distribution_performance}
\end{figure*}

\subsection{Computational Implications}
Table \ref{tab:efficiency} demonstrates the bottleneck of computation for \ours{}. Specifically, The first row calculates the average LLM calls conducted during the Table Augmentation with Docs References and Term Explanations, which contributes to $N + C$ times LLM calling. Here, $N$ is the number of terms detected by the LLM, and $C$ is a constant equal to 2 if there is no error occurs when calling the API. 
We also measure the token usage for each LLM call during the Table Sampling. Since the Table Sampling module only retrieves the top-k-related rows/columns to reconstruct a sub-table for LLM calling, this ensures token expenditure without compromising the model's performance, especially in cases where thousands of rows exist in a table. The second row of the table \ref{tab:efficiency} shows the average token usage per question needed by TAP4LLM after table sampling.
\begin{table}[H]
  \centering
  \caption{\small{Efficiency of multiple preprocessing steps in TAP4LLM}}
  \resizebox{1\linewidth}{!}{
    \begin{tabular}{lcccc}
    \toprule
    Dataset & SQA & FEVEROUS & TabFact & HybridQA \\
    \midrule
    Average LLM Calls & 4.7 & 5.2 & 5.3 & 8.4 \\ 
    Average Token Usage & 637 & 512 & 417 & 742 \\ 
    \bottomrule
    \end{tabular}%
  }
  \label{tab:efficiency}%
\end{table}

\subsection{Large Table Analysis in \ours{}}
Compared to smaller tables, large tables can grow to immense sizes, making them more difficult to maintain and reason over tabular data. 
In designing \ours{}, performance optimization in this context is essential.
Figure~\ref{fig:distribution_performance} presents the distribution of token counts from the table across the five datasets, while also illustrating the impact of token numbers on LLM performance in three distinct settings.
We can observe that:
(1) Shorter token lengths dominate the datasets, indicating the prevalence of text entries is relatively brief. (2) Augmentation techniques excel with these shorter lengths by providing focused, enriched contexts that facilitate better model learning from simpler inputs. In contrast, sampling methods prove more effective for larger tables, suggesting that they help manage data complexity by focusing on relevant data segments. (2) The hybrid method shows consistent performance across various token lengths, highlighting its ability to leverage the strengths of both augmentation and sampling for robust performance enhancement.
\section{Related Work}
\label{sec:related_work}

\paragraph{Large Language Models for Tabular Data} Following the line of LLMs in natural language processing, researchers have also explored large models for various modalities like vision~\cite{gong2023multimodal,kirillov2023segment} and speech~\cite{huang2023audiogpt}. From a technical standpoint, their ability to generate human-like text has opened new vistas of possibilities for processing tabular data. Nevertheless, it is non-trivial to directly employ the vanilla LLMs in the tabular area for two reasons: (\romannumeral1)-Global Table Understanding: the GPTs are known to suffer from the limited token length and thus, they can not read a whole large table, making them hard to understand the global tabular information. (\romannumeral2)-Generalized to Tabular Domain: Second, their training processes are tailored for natural languages and thus, they are less generalizable when handling tabular data. There have been several works \cite{hu2023chatdb,zhong2017seq2sql,li2023nl2sql,li2023sheetcopilot} developed to integrate natural language for tabular data analysis. 

\paragraph{Table Augmentation} Table augmentation is a technique used to improve the generalization performance and robustness of machine learning models.
To enhance the performance and capabilities of LLMs in various domains, various explorations have been done to augment their knowledge grounding. It involves incorporating structured knowledge~\cite{model_structure_LLM,model_unifiedSKG}, commonsense knowledge~\cite{model_chatgpt_commonsense,model_chatgpt_analysis,model_ischatgpt_trust,model_how_chatgpt_close_human}, and analytical knowledge~\cite{he2023anameta,model_table_recasting} into the pre-training and inference processes. 
For example, \cite{model_table_recasting} proposes to semi-automatically transform existing tabular data to create diverse/inventive natural language inference instances for better zero-shot performance. 
\cite{he2023anameta} proposes a multi-tasking Metadata model that leverages field distribution and knowledge graph information to accurately infer analysis metadata for tables, and then demonstrates its deployment in a data analysis product for intelligent features. 
\section{Conclusion}
\label{sec:conclusion}


In this paper, we propose \ours{} (Table Provider for LLM) as a powerful toolkit designed to enhance the interaction between LLMs and structured table data. It provides optimized prompt designs and robust functionalities to ensure high-quality outputs when LLMs process table-related inputs. We believe that \ours{} has the potential to significantly improve table modeling and exploratory data analysis (EDA), with applications in various domains. 


\section*{Limitations}
\label{sec:limitations}

Code generation-based methods~\cite{model_binder, model_tool_synthesis, he2024conline} have been proposed to leverage LLMs to convert natural language queries into executable code or structured representations. We believe that semantic parsing or code generation is an important research direction. However, due to the page limits, we will leave this topic to further exploration.
Additionally, our empirical study is mostly designed for English, rather than multilingual scenarios. The conversation on multilingual capabilities will also be part of future exploration.

\section*{Ethics Statement}
\label{sec:ethics}

All the datasets used in this paper are public and have been reviewed to ensure they do not contain any personally identifiable information or offensive content. However, as these datasets are sourced from the Internet, potential bias may still be present. Furthermore, despite our careful review, the process of table augmentation with LLMs throughout may inadvertently introduce inappropriate information into the preprocessed data. In addition, all the experiments in this paper are run on GPU clusters with 8 NVIDIA A100 GPUs. 


\bibliography{references,references_zotero}

\appendix
\label{sec:appendix}

\section{Rule-based Sampling}
\label{sec:rule-based-sampling}

Rule-based sampling refers to table sampling based on predefined criteria or rules. These methods follow the established patterns or criteria for data selection. We consider three common rule-based sampling methods as follows: (1) \textit{Random Sampling}, by selecting rows from a table, with each having an equal probability of being selected. To increase the quality of this baseline, we repeat the random selection for a user-specified amount of time and return the sub-table with the highest combined score among all the randomly computed sub-tables.
(2) \textit{Evenly Sampling}:
It samples rows from a table by alternating between the top ($r_1$) and bottom rows ($r_n$) and moving towards the middle until reaching a set token limit. Compared to random sampling, it helps to balance the proportions of each field in the dimension column of the table (\textit{i.e.} rows are selected at regular intervals), ensuring a uniform distribution of the sample accross the entire table.
(3) \textit{Content Snapshot \& Synthetically Sampling}: Content snapshot~\cite{model_tabert} is a text-matching based method for retrieving sub-tables. For our empirical analysis,
we construct the content snapshot $K$ rows based on their relevance to the utterance using $n$-gram overlap ratio.
Specifically, for $K > 1$, top-$K$ rows with the highest $n$-gram overlap ratio are selected. For $K = 1$, a synthetic row is composed by selecting the cell values from each column with the highest $n$-gram overlap with the utterance. 
The comparative results can be found in Table~\ref{tab:table_sampling}.

\section{Retrieval-based Augmentation}
\label{sec:retrieval-based-augmentation}

\subsection{Docs References}
This process involves associating tables with relevant documents or sources for in-depth insights or references. 
For example, suppose we have a table titled ``2023 Fortune 500 Companies''. This table contains various information about the top 500 companies as ranked by Fortune in 2023, including their revenue, number of employees, and market capitalization. Docs references could fetch the actual 2023 Fortune 500 list from the Fortune website, Wikipedia pages discussing the Fortune 500 concept and its criteria, or analytical articles discussing the companies on the 2023 list.
In our setting, we leverage Langchain~\cite{langchain} to retrieve wiki pages from wikipedia.org. We craft queries by concatenating the table header and the table's title into a single string. These queries are then used to identify and fetch the relevant Wikipedia pages, which act as informative document references in our study.

\subsection{Term Explanation}
Compared to the docs references, term explanation focuses on providing definitions and explanations for specific strange terms or values in the table cells. For example, if a cell mentions a technical term or an acronym, the term explanation module could source a brief definition or background from reliable web sources (such as Wikipedia, wolfram,\etc) on that term, ensuring that the strange term will not be forwarded to LLMs. To ensure the efficacy and accuracy of term explanations, we introduce two distinct approaches for selecting the cell that is required to be explained, \textit{LLM-based Cell Selection} and \textit{Heuristics-based Cell Selection}. The comparative experiment results of these two variants can be found in Table~\ref{tab:table_augmentation}.

\textit{1) LLM-based Cell Selection Module}:
To pinpoint the exact cell warranting explanation, we harness the capabilities of LLMs. The selection prompt is meticulously constructed, taking into account various factors including:
(1) Cell Position;
(2) Cell Content;
(3) Cell Formatting;
(4) Cell Context;
(5) Cell Properties.
A detailed description and the specific prompt utilized to determine which cells require explanation can be found in Table~\ref{tab:prompt}.

\begin{table}[htbp]
  \centering
  \caption{\small{LLM-based Cell Selection Criteria and Exact Prompt Template.}}
  \label{tab:prompt}
    \resizebox{\linewidth}{!}{
    \begin{tabular}{p{6.5em}p{21.835em}}
    \toprule
    Criteria & Description \\
    \midrule
    Cell Position & Specify the range or position of the cells you want to search. For example, you may want to search for explanations only in the cells of a specific column, row, or a particular section of the table. \\
    \midrule
    Cell Content & Define the specific content or data type within the cells you want to search. For instance, you may want to search for explanations in cells containing numerical values, dates, specific keywords, or a combination of certain words. \\
    \midrule
    Cell Formatting & Consider the formatting or styling applied to the cells. This could include searching for explanations in cells with bold or italic text, specific background colors, or cells that are merged or highlighted in a certain way. \\
    \midrule
    Cell Context & Take into account the context surrounding the cells. You can search for explanations in cells that are adjacent to certain labels, headings, or identifiers, or within a specific context provided by other cells in the same row or column. \\
    \midrule
    Cell Properties & Consider any specific properties associated with the cells. This might include searching for explanations in cells that have formulas, links, or other data validation rules applied to them. \\
    \midrule
    \textbf{Prompt} & \cellcolor[rgb]{ .847,  .882,  .953}You will be given a parsed table \textbf{\{Table\}} in python dictionary format, extract the cells that need to be explained. The extraction rule should be based on the following criteria: \textbf{\{Criteria\}}. Only return the cells name in a python List[str]. \\
    \bottomrule
    \end{tabular}}
\end{table}

\textit{2) Heuristics-based Cell Selection}:
Inspired by the methodology presented in \cite{model_tapas}, we introduce a heuristics-based cell selection, which is predicated upon the following criteria:
(1) Explicit Mention: whether the cell’s value is explicitly referenced in the query.
(2) Comparative Value: whether the cell’s value is greater or less than a value mentioned in the query.
(3) Superlative Value: whether the cell’s value represents a maximum or minimum across the entire column, especially when the query incorporates superlative terms.

\section{Metadata-based Augmentation}
\label{apx:metadata_type}

Metadata are defined as a form of formally represented background knowledge to understand the field semantics for correctly operating on table fields (or columns) and to further find common patterns in daily analysis~\cite{he2023anameta}. This analytical knowledge, particularly of field semantics, is able to increase the applicability across various tasks. In our table augmentation, we consider the following metadata:

(1) \textit{Dimension / Measure}: This is one type of metadata used in Tableau~\cite{hoelscher2018} and Excel~\cite{ding2019a} across diverse features. As the name suggests, the method involves categorizing each field in a table as either measure or dimension. 
The measure contains numerical data that can be subjected to calculations, such as the ``Price" and ``Discount". The dimension provides categorical information used for filtering, grouping, and labeling, such as the ``Product Name" and ``Category". Correctly classifying fields as either a measure or a dimension is crucial to determining feasible operations on the data and influences the accuracy and relevance of data analysis.
(2) \textit{Semantic Field Type}: 
Besides identifying whether a field is a measure or a dimension, semantic field type specifies the meaning and format of the data within each field based on knowledge graphs. For example, the dimension field includes semantic field types such as ``Consumer Product'' and ``Category'', \etc. Measure field includes semantic field types such as ``Money'' and ``Ratio'', etc. We follow the work ~\cite{he2023anameta} as a reference to this term. 
(3) \textit{Table Size}: The size of a table is defined by its number of rows and columns. It provides essential context when determining the computational complexity of operations or understanding data density and granularity. 
(4) \textit{Statistics Feature}: Statistics feature provides a quantitative representation of the tabular data. These features serve as numerical descriptors that summarize key aspects of the table datasets, aiding LLMs in understanding the overall characteristics and tendencies. Generally, statistics features include four categories~\cite{he2023anameta}: (a) Progression features (b) String features (c) Number range features (d) Distribution features, discussed in Section~\refsec{sec:related_work}. We conducted empirical studies on common statistical features to identify the most appropriate combination for optimal utilization of ~\ours{}.
(5) \textit{Header Hierarchy}: Tables are often used to present data in a structured format, and headers play a crucial role in defining the meaning and context of the data in each column or row. The header hierarchy typically includes different levels of headers, each providing a level of organization and categorization for the data.

\section{Additional Experiment Settings}
\label{sec:experiment_settings}

\subsection{Downstream Tasks and Datasets}
\label{sec:tasks_and_datasets}
\textbf{Table Reasoning Tasks}.
Each instance in table-based reasoning consists of a table $T$, a natural language question $Q$, and an answer $A$. Specifically, table $T$ is defined as $T=\left\{v_{i, j} \mid i\leq\operatorname{R}_T, j\leq\operatorname{C}_T\right\}$, containing $R_T$ rows and $C_T$ columns. The content of the cell in the $i$-th row and $j$-th column is represented by $v{i, j}$. A question $Q$ is a sequence of $n$ tokens: $Q=\{q_1, q_2, q_3, \cdots, q_n\}$.
In this paper, our primary focus is on two distinct table-based reasoning tasks, table-based fact verification (TFV) and table-based question answering (TQA). In TFV, the answer $A$ is a boolean value in $\{0,1\}$, indicating the veracity of the input statement (where $1$ means the statement is entailed by the given table, and $0$ means the statement is refuted by the given table). In TQA, the answer is a sequence of natural language tokens represented as $A=\{a_1, a_2, a_3, \cdots, a_n\}$ corresponding to the posed question.
For our experiments, all tables first undergo table sampling and table augmentation by our proposed method and then are serialized into a sequence by table packing and serialization. Detailed implementation specifics are provided in Section~\refsec{sec:table_packing_serialization}.
\begin{table*}[htbp]
  \centering
  \caption{\small{The distribution of the used datasets.}}
  \label{tab:dataset_distribution}
  \resizebox{\linewidth}{!}{
  \begin{tabular}{lcccccc}
    \toprule
    Property & SQA & FEVEROUS & TabFact & HybridQA & ToTTo & Spider\\
    \midrule
    Unique Query (Set Size) & 1,228 & 1,322 & 9,228 & 6,268 & 8,026 & 10,181\\
    Unique Table & 432 & 942 & 1,342 & 4,364 & 5,934 & 500\\
    SQL Query & - & - & - & - & - & 5,693\\
    Rows per tables (Median/Avg) & 12 / 18.5 & 14 / 26.3 & 8 / 14.0 & 8 / 15.7 & 16 / 28.4 & 10 / 16.1\\
    Columns per tables (Median/Avg) & 4 / 6.4 & 4 / 5.5 & 4 / 5.5 & 4 / 4.3 & 6 / 8.8 & 4 / 4.5\\
    Cells per tables (Median/Avg) & 78 / 180.4 & 77 / 190.3 & 80 / 150.3 & 70 / 143.9 & 87 / 212.6 & -\\
    \midrule
    Domain & Wikipedia & Wikipedia & Wikipedia & Wikipedia & Wikipedia & -\\
    Evaluation Metric & Exact Match & Exact Match & Exact Match & Exact Match & BLEU-4 & Execution Accuracy\\
    \bottomrule
  \end{tabular}}
\end{table*}
\begin{table*}[htbp]
  \centering
  \caption{\small{Different kinds of table augmentation.}}
  \label{tab:table_augmentation_definition}
  \resizebox{\linewidth}{!}{
    \begin{tabular}{p{13.335em}p{14em}p{31.335em}}
    \toprule
    \textbf{Knowledge Aspect} & \textbf{Categories} & \textbf{Definition} \\
    \midrule
    Dimension/Measure & Metadata-based & Distinguish each element in a table as either dimension field or measure field.\\
    \midrule
    Semantic Field Type & Metadata-based & Classify the meaning and format of the data within each field based on knowledge graphs.\\
    \midrule
    Table Size & Metadata-based & Basic information of a table including numbers of rows and columns.\\
    \midrule
    Statistics Feature & Metadata-based & Statistics features such as change rate, numerical distribution, range of data.\\
    \midrule
    Header Hierarchy & Metadata-based & The organization and structure of header elements within a table. \\
    \midrule
    Docs References & Retrieval-based & External domain knowledge from reliable webpages (\eg, wikipedia, Wolfram Alpha, \etc.) which are similar to the given context. \\
    \midrule
    Term Explanation & Retrieval-based & External domain knowledge such as term and metric definitions (formulas, relevant documents/sources, search results, etc.) \\
    \midrule
    Self Prompting & Self-consistency-based & Leverage LLMs to generate some reasoning thoughts as supplementary for table augmentation (self-augmented prompting, chain-of-thoughts, \etc.) \\
    \bottomrule
    \end{tabular}
    }
\end{table*}

In this paper, we mainly focus on tabular reasoning with two major tasks: TQA \& TFV. We conduct experiments on five typical datasets and the distribution of the datasets can be found in Table~\ref{tab:dataset_distribution}. In addition, to extend our work to databases containing table structures, we also set up ~\ours{} on Spider ~\cite{yu2018spider} dataset.
Specifically, we use: 
(1) \textbf{SQA}~\cite{dataset_sqa}, which is constructed by decomposing a subset of a highly compositional dataset, WTQ~\cite{dataset_wikitq}. The dataset consists of 1,288 unique queries corresponding to 432 tables, with each table having 18.5 rows and 6.4 columns on average; 
(2) \textbf{HybridQA}~\cite{dataset_hybridqa}, which is designed as a large-scale multi-hop question-answering dataset over heterogeneous information of both structured tabular and unstructured textual forms. The dataset consists of 6,268 unique questions and each question is aligned with a Wikipedia table. Compared to the SQA dataset, HybridQA has shorter column numbers, which facilitates the understanding of the table's structure boundaries.
(3) \textbf{ToTTo}~\cite{dataset_totto} is a high-quality English table-to-text dataset. 
It proposes a controlled generation task that involves synthesizing a one-sentence description given a Wikipedia table and a set of highlighted table cells. The dataset contains 8,026 samples, each comprising a Wikipedia table with highlighted cells. Each table contains 16 rows and 6 columns on average.
(4) \textbf{FEVEROUS}~\cite{dataset_feverous} is a fact verification dataset over structured information. The dataset consists of 1,322 verified claims. Each claim is annotated with evidence in the form of sentences and cells from tables in Wikipedia. Each annotation also includes a label indicating whether the evidence supports, refutes, or does not provide enough information to make a decision. Each table contains 26.3 rows and 5.5 columns on average.
(5) \textbf{TabFact}~\cite{dataset_tabfact} is another fact verification dataset where the tables are extracted from Wikipedia and the sentences are composed by crowd workers. Compared to the FEVEROUS dataset, TabFact encompasses a larger number of samples and each table has fewer rows, has 14 rows per table on average.

\textbf{Metrics.}
For TQA and TFV tasks (SQA, FEVEROUS, TabFact and HybridQA), we report the exact match accuracy of answer sets. For the data-to-text generation task (ToTTo), we report the BLEU-4 score. 

\subsection{Models}
\label{sec:models}
We evaluate the performance of the recent dominant LLM models, 1) Instruct-GPT-3.5~\cite{model_instruct_gpt3}, using versions gpt-3.5-turbo, gpt-3.5-turbo-16k; 2) GPT-4, using the latest version of gpt-4 model; 3) Llama-2-70B ~\cite{touvron2023llama}, using version 17; 4) Mixtral-8x7B ~\cite{jiang2024mixtral}, using version 0.1. 

Unless otherwise specified, we utilize \textbf{gpt-3.5-turbo} in all experiments. In the sampling methods, we use text-embedding-ada-002~\cite{openai_embedding} for row and query embedding generation. 
The comparison experiments using other embeddings models, such as, text-search-ada-doc-001, bge-largen-en~\cite{bge_embedding}, all-MinLM-L6-v2~\cite{reimers2019sentence} can be found in Table~\ref{tab:embedding_methods}. We set the temperature of all the models as 0, top p as 1.0, frequency penalty as 0, and presence penalty as 0.

The development of \ours{} begins with the foundation provided by LLMs. In designing our framework, we opt to use OpenAI models as our base model due to their excellent capabilities in language reasoning. However, the choice is not exclusive. Since \ours{} use natural language as an intermediary for interactive communication between the table and LLMs, it can also support other outstanding open-sourced models using natural language as input, such as  Phoenix~\cite{chen2023phoenix}, ChatGLM~\cite{zeng2022glm}, Ziya~\cite{Fengshenbang-LM}, and Baichuan~\cite{baichuan}. This design provides versatility and flexibility in \ours{} implementation.

\section{Additional Experiments}
\label{sec: additional_experiments}

\subsection{Comparison Results of Embedding Type.}
Based on the results from Table~\ref{tab:embedding_methods}, we observe that:
(1) \textit{Superiority of ``text-embedding-ada-002''}: ``text-embedding-ada-002'' consistently offers the best performance across the datasets. It suggests that for tasks similar to table reasoning, this embedding type might be the most suitable choice.
(2) \textit{Potential of ``sentence-transformer''}: The ``sentence-transformer'' embedding type provides competitive results, especially in the ToTTo dataset. This suggests that it might be particularly suitable for certain tasks or datasets and is worth considering alongside ``text-embedding-ada-002''.

\begin{table}[htbp]
  \centering
  \caption{\small{Comparative results of different embedding models on query-based sampling method without any augmentation method. We use all-MinLM-L6-v2 for the sentence-transformer. The highest performance of each dataset is bold.}}
  \label{tab:embedding_methods}
  \resizebox{1\linewidth}{!}{
    \begin{tabular}{lcccccc}
    \toprule
    Embedding Type & SQA & FEVEROUS & TabFact & HybridQA & ToTTo & Spider\\
    \midrule
    text-embedding-ada-002 & \textbf{28.32\%} & \textbf{63.32\%} & \textbf{59.80\%} & \textbf{24.32\%} & 49.14\% & \textbf{80.27\%}\\
    text-embedding-ada-001 & 27.12\% & 62.24\% & 57.32\% & 23.14\% & 48.21\% & 79.34\%\\
    bge-large-en~\cite{bge_embedding} & 26.76\% & 62.87\% & 56.31\% & 22.65\% & 47.32\% & 78.25\%\\
    sentence-transformer~\cite{reimers2019sentence} & 26.32\% & 63.31\% & 58.94\% & 23.78\% & \textbf{50.12\%} & 80.05\%\\
    \bottomrule
    \end{tabular}}
\end{table}


While ``text-embedding-ada-001'' and ``bge-large-en'' don't lead to the highest performance, they still provide competitive performance. This suggests that the choice of embedding can affect the overall performance, but the differences might not always be significant. The choice between these embeddings would likely depend on specific use cases, computational costs, and other practical considerations.

\subsection{Comparison Results of Statistics Features}
The accuracy of each dataset for four groups of statistics features reveals that the distribution features overall performed well in capturing the nuances and variations within specific tabular data entries. Based on this, we further propose a combination including the most practical features across these four categories and carry out an empirical study to examine its performance. Specifically, this combination contains variance, range, cardinality, major, and change rate. with each term's definition listed in Table~\ref{tab:statistics_features}. The experiment result, displayed in Table~\ref{tab:statistics_feature_results}, demonstrates that our proposed combination surpasses the previous four feature sets across all six datasets.
\begin{table}[htbp]
  \centering
  \caption{\small{Detailed definition of statistics features.}}
  \label{tab:statistics_features}
  \resizebox{1\linewidth}{!}{
    \begin{tabular}{ll}
    \toprule
    \textbf{Features} & \textbf{Definition}\\
    \midrule
    \textbf{Progression Type:} \\
    ChangeRate & Proportion of different adjacent values \\
    PartialOrdered & Maximum proportion of increasing / decreasing adjacent values\\
    OrderedConfidence & Indicator of sequentiality \\\\
    \textbf{String Features:} \\
    AggrPercentFormatted & Proportion of cells having percent format \\
    CommonPrefix & Proportion of most common prefix digit\\
    CommonSuffix & Proportion of most common suffix digit\\\\
    \textbf{Number Range Features:}\\
    Aggr01Ranged & Proportion of values ranged in 0-1 \\
    Aggr0100Ranged & Proportion of values ranged in 0-100 \\
    AggrIntegers & Proportion of integer values \\
    AggrNegative & Proportion of negative values\\\\
    \textbf{Distribution features: } \\
    Variance & Standard deviation of a given series of data \\
    Range & Values range \\
    Cardinality & Proportion of distinct values \\
    Spread & Cardinality divided by range \\
    Major & Proportion of the most frequent value\\
    Benford & Distance of the first digit distribution to real-life average\\
    Skewness & Skewness of numeric values \\
    Kurtosis & Kurtosis of numeric values \\
    Gini & Gini coefficient of numeric values \\
    \bottomrule
    \end{tabular}}
\end{table}

\begin{table}[htbp]
  \centering
  \caption{\small{Comparative results of various types of statistical features. The experiment setting is the same as Section~\ref{tab:table_augmentation}. The highest performance of each dataset is bold.}}
  \label{tab:statistics_feature_results}
  \resizebox{1\linewidth}{!}{
    \begin{tabular}{lccccccc}
    \toprule
    Statistics Features Type & SQA  & FEVEROUS & TabFact & HybridQA & ToTTo & Spider\\
    \midrule
    Progression features            & 29.20\%    &  64.26\%   & 60.45\%  & 25.11\%  & 49.53\%   & 77.47\%\\
    String features                 & 28.56\%    &  63.13\%   & 61.38\%  & 24.83\%  & 48.29\%	& 73.56\%\\
    Number range features           & 29.13\%    &  62.18\%   & 59.03\%  & 24.53\%  & 49.68\%	& 76.32\%\\
    Distribution features           & 30.28\%    &  66.34\%   & 62.18\%  & 24.76\%  & 49.34\%	& 79.14\%\\
    Statistics features  & \textbf{31.22\%}    & \textbf{66.51\%}   & \textbf{62.33\%}  & \textbf{26.13\%}  & \textbf{50.57\%}  & \textbf{80.94\%}  \\
    \bottomrule
    \end{tabular}}
\end{table}

\subsection{\ours{} in Open-source model} 
\label{sec: open_source}
Beyond conducting experiments on GPT models. we also evaluate the effectiveness of \ours{} on two recent LLMs: Llama-2-70B and Mixtral-8x7B. According to Table~\ref{tab:table_opensource}, we first evaluated direct inference on open-source models and then apply \ours{} to each model. The result demonstrates that \ours{} increases models' performance on all five datasets. 

In addition, \ours{} is currently evaluated under the setting of in-context learning. However, through parameter-efficient fine-tuning \cite{hu2021lora} and recently advanced prompt compression techniques \cite{jiang2023llmlingua, zou2024promptintern}, we can directly apply \ours{} on more challenging tabular reasoning tasks requiring training procedures. We will leave this as our future work. 
\begin{table}[htbp]
  \centering
  \caption{\small{Comparison of TAP4LLM and baseline on Open-source models. We refer to "Baseline" as directly inferring each task using the model. For TAP4LLM, we apply semantic sampling for table sampling module and Statistics Feature/D/M+SF/self-prompting for table augmentation module.}
}
  \label{tab:table_opensource}
  \resizebox{\linewidth}{!}{
    \begin{tabular}{llccccc}
    \toprule
     Model Name & \multicolumn{1}{l}{Methods} & SQA & FEVEROUS & TabFact  & HybridQA & ToTTo\\
    \midrule
    \multirow{2}{*}{Llama-2-70B} & Baseline &  19.02\% & 65.33\% & 63.45\% & 17.21\% & 21.08\% \\
                                 & TAP4LLM  &  22.14\% & 69.20\% & 66.32\% & 23.15\% & 30.00\% \\
    \midrule
    \multirow{2}{*}{Mixtral-8x7B}& Baseline &  21.25\% & 61.32\% & 57.21\% & 21.01\% & 34.25\% \\
                                 & TAP4LLM  &  24.18\% & 63.29\% & 58.80\% & 25.44\% & 37.79\% \\
    \bottomrule
    \end{tabular}
    }
\end{table}%

\subsection{\ours{} in Database Application} \label{sec: spider}
\textbf{Dataset} We test \ours{} effectiveness on Spider~\cite{yu2018spider}. Spider is a cross-domain Text-to-SQL dataset as shown in Table \ref{tab:dataset_distribution}. Each instance contains a natural language question, a specific database containing tabular information, and one corresponding SQL query.\\
\textbf{Metric} We evaluate \ours{} on the development split \textit{Spider-dev} which contains 1034 instances over 200 databases. We use the Execution Accuracy, followed by the original paper~\cite{yu2018spider}, to compare the execution output of the predicted SQL query with the golden SQL query. \\
\textbf{Experiment} As shown in Table~\ref{tab:table_spider_sampling} and Table~\ref{tab:table_spider_augmentation}, the experiment result demonstrates that LLMs achieve an overall higher model performance through \ours{}. Specifically, the execution accuracy reaches the highest through semantic-based sampling and D/M + SF augmentation. 

\begin{table}[htbp]
  \centering
  \caption{\small{Comparative results of the table sampling methods on Spider.}
}
  \label{tab:table_spider_sampling}
  \resizebox{\linewidth}{!}{
    \begin{tabular}{llc}
    \toprule
     \textbf{Sampling Type} & \multicolumn{1}{l}{Table Sampling Methods} & Execution Accuracy \\
    \midrule
    \multirow{3}{*}{Rule-based Sampling} & Random Sampling & 74.58\%\\
    & Evenly Sampling & 72.03\%\\
    & Content Snapshot~\cite{model_tabert} & 78.93\% \\
    \midrule
    \multirow{4}{*}{Embedding-based Sampling} &
    Centroid-based Sampling & 77.43\% \\
    
    & Semantic-based Sampling & 80.27\% \\
    
    & \quad w/ Column Grounding  & \textbf{81.03}\%\\
    
    & Hybrid Sampling & 78.94\% \\
    \midrule
    \multirow{1}{*}{LLM-based Sampling} & LLM-Decomposer~\cite{ye2023} & 78.34\% \\
    \midrule
    \multirow{2}{*}{-} & No sampling (GPT-3.5) & 72.15\% \\
    
    & No sampling (GPT-3.5, truncated)  & 68.47\% \\
    \bottomrule
    \end{tabular}
    }
\end{table}%

\begin{table}[htbp]
    \label{tab:spider_augmentation}
  \centering
  \caption{\small{Comparative results of table augmentation methods on Spider. We use semantic-based sampling method without augmentation as the default method for table augmentation.}
}
  \label{tab:table_spider_augmentation}
  \resizebox{\linewidth}{!}{
    \begin{tabular}{lc}
    \toprule
    \textbf{Augmentation Aspect} & Execution Accuracy\\
    \midrule
    Baseline & 80.27\%\\
    \midrule
    D/M + SF & \textbf{82.45}\% \\
    
    Statistic Feature & 80.94\% \\
    
    Term Explaination (LLM-based)  & 80.48\% \\
    
    Term Explaination (Heuristics-based) &80.33\% \\
    
    \bottomrule
    \end{tabular}
    }
\end{table}%

\section{Implementation Details}
\label{sec:discussion}

\subsection{Motivation of our Framework}

\textbf{Table Sampling}: One primary challenges for tabular reasoning is that the full content of a table could be very long and noisy to be include in the prompt. Most LLMs have a limited input context window size (\eg, 4k tokens) in which an overlong table cannot fit it. For long tables that satisfy the length constraint, it can still lead to unnecessary computations (of LLMs on long prompt) and quality regressions (generation interfered by noisy input) when placing irrelevant table content (\textit{w.r.t.} the task or query) in the prompt.\\ 
\textbf{Table Augmentation}: Another challenge is what additional/external knowledge could help LLMs better understand a table? The raw content of a table may contain ambiguous information (\eg, abbreviations, domain-specific terms, column type, \etc) that requires further interpretation and clarification. We are motivated to propose table augmentation for 1) \textit{enhanced contextual understanding}: by supplementing tables with metadata and attributes, we can achieve a more profound grasp of the table's intrinsic structure and semantics and further enrich the tabular data; 2) \textit{bridging external knowledge gasps}: tables alone might not encompass all the required information to provide comprehensive answers to certain queries. By retrieving external knowledge from reliable sources, \eg, Wikipedia, we can aid the language models in understanding the broader context of the query, leading to more informed and nuanced responses.\\
\textbf{Table Packing}: The desire to maintain efficient reasoning without changing the LLMs architecture motivates us to consider how to encode the table into a prompt? While sampling and grounding compress the table content, augmentation expands the prompt by adding more information. With a given token budget, one needs to find the balance to allocate available tokens between table content and augmented knowledge.

\subsection{Table Syncing}
\label{sec:interactive_table_reasoning}
To achieve the interactive table reasoning,
\ours{} proposes the ``table sync'' to ensure that applications, such as Excel Copilot, maintain their table data in synchronization with the table manager. The table manager acts as a go-between, managing the data that is either stored locally in a cache or accessed through a database connection. Specifically, when changes are made to the data within the application, those changes must be reflected in the table manager for any operation performance, such as sampling, augmentation, and packing. Conversely, if changes are made within the table manager, the changed data should be updated in the application as well.

This syncing process is essential for maintaining data integrity and ensuring that all components of the system are kept up-to-date. This is especially beneficial when the data is being used to generate prompts for a large language model, as it allows for accurate data processing, querying, and analysis. By having the most current and relevant information, the model can provide accurate and reliable responses.

\subsection{Table Cleansing}
\label{sec:table_cleansing}

Table cleansing is an independent step in tabular data prepossessing, especially when dealing with hierarchical tables. In the context of fine-grained in-context learning, where pre-trained generated model has to discern and process intricate patterns and relationships within datasets. The importance of clean and standardized tables cannot be overstated for two reasons: (1) Dirty or unorganized tabular data can mislead the models and impair the model's performance; (2) Cleansed tables ensure uniformity, making them easier to compare, merge, or use in subsequent operations. For example, imagine a financial analyst case aiming to forecast a company's stock price based on historical data. The corresponding table contains daily stock prices, trading volumes, and various financial indicators. If there are any missing certain values for certain days, or duplicate entries due to system glitches. Such inconsistencies may dramatically affect the forecasting performance. For instance, it might suggest a non-trading day or a sudden drop in stock price.
Specially, the formal definition of table cleansing is: Given a table $T$ consisting of rows $R_T$ and columns $C_T$, table cleansing transforms $T$ into $T'$ such that:
(a) \textit{Cell and column name completeness}: For every cell $c_{i,j}$ in $T$ where $i \in R_T$ and $j \in C_T$, if $c_{i, j}$ has a missing or null value, it is filled using contextual information (\ie, use the corresponding entire column $C_j$ of cell $c_{i,j}$ as the context). We utilize a separate ``CallLLM" system $g(\cdot)$ to call a pre-trained language model to synthesize the missing value. The processing can be formulated as $c_{i,j}=g(C_j)$. This ensures that gaps in the data don't lead to misleading interpretations or missed patterns.
(b) \textit{Duplicate data points removal}: For every pair of rows $r_m$, $r_n$ and pair of columns $c_p$, $c_q$ in $T$, if $r_m=r_n$ or $c_p=c_q$ respectively, one from the pair is removed to eliminate duplication.
(c) \textit{Format consistency}: For every cell $c_{i,j}$ in $T$, the value conforms to a specific format, unit, or pattern.

\end{document}